\documentclass[letterpaper, 10 pt, conference]{ieeeconf}  

\IEEEoverridecommandlockouts                              

\author{Bonan Shen$^{1}$, Wei-Jung Huang$^{1}$, Xin Liu$^{1}$, Jiazhou Gao$^{1}$ and Tao Ning$^{1}$ 
\thanks{$^{1}$Independent Researcher \{\small shenbonan2, william.wj.huang, iamxinliu, gjz140103, ntgd1102\}@gmail.com}
}

\usepackage{amsmath,amssymb,amsfonts}
\usepackage{algorithmic}
\usepackage{algorithm}
\usepackage{graphicx}
\usepackage{textcomp}
\usepackage{xcolor}
\usepackage{booktabs}
\usepackage{multirow}
\usepackage{tikz}
\usepackage{pgfplots}
\pgfplotsset{compat=1.18}
\usetikzlibrary{patterns,calc,positioning}
\usepackage{float}
\usepackage[colorlinks=true,linkcolor=blue,citecolor=blue,urlcolor=blue]{hyperref}
\usepackage{array}

\begin{document}
\hbadness=10000

\title{AdaStop: Cost-Aware Early Stopping for DNN Test Selection}


\maketitle

\begin{abstract}

Existing methods for testing deep neural networks (DNNs) primarily prioritize test inputs likely to reveal model faults under a fixed labeling budget. In practice, choosing that budget is difficult: too little testing misses failures, while too much incurs unnecessary labeling costs. This work studies the stopping problem in DNN testing. We formulate testing as a cost--benefit decision process in which labeling an input incurs cost $c$ and discovering a fault yields value $v$. Based on this formulation, we introduce \textit{AdaStop}, a framework that estimates the marginal fault discovery rate during testing and stops labeling when the estimated rate falls below the threshold $\tau = c/v$. Experiments across multiple datasets, architectures, and selection strategies show that $65$--$84\%$ of faults can be discovered using only $9$--$31\%$ of the labeling budget.

\end{abstract}

\section{Introduction}

Deep neural networks (DNNs) are increasingly deployed in domains where reliability is critical, including autonomous vehicles~\cite{tian2018autonomouscars}, medical diagnosis~\cite{esteva2017skincancer}, and financial systems~\cite{ozbayoglu2020financial}. Ensuring reliability requires systematic testing to identify test inputs on which the model makes incorrect predictions---commonly referred to as \textit{faults}. In many settings, however, verifying predictions requires ground-truth labels obtained through costly human annotation by domain experts~\cite{settles2010activelearning}.

This creates a fundamental and practical challenge: \textit{when should we stop labeling test inputs?} Stopping too early risks missing faults that could lead to system failures; stopping too late wastes labeling resources on inputs that are unlikely to reveal new faults.

Existing test selection methods primarily focus on the \textit{selection} problem---determining which inputs to label first---while the complementary \textit{stopping} problem has received comparatively little attention. We identify three key limitations:
\begin{enumerate}
    \item \textbf{Lack of principled stopping criterion.} Methods like DeepGini~\cite{feng2020deepgini}, TestRank~\cite{li2021testrank}, ATS~\cite{gao2022ats}, and DeepSample~\cite{guerriero2024deepsample} typically evaluate performance under predetermined labeling budgets without providing guidance on when testing should terminate.
    \item \textbf{Arbitrary budget selection.} In the absence of stopping criteria, practitioners must choose labeling budgets heuristically. However, appropriate budgets depend on factors such as model quality and dataset characteristics, which are often unknown beforehand.
    \item \textbf{Unaddressed cost--benefit trade-off.} Labeling each input incurs cost, while discovering a fault provides value. Existing approaches focus on optimizing the \textit{order} of test selection, such as prioritizing inputs that are more likely to reveal model errors, but do not explicitly consider this cost--benefit trade-off~\cite{feng2020deepgini}.
\end{enumerate}

Although stopping criteria exist in related domains, such as the SAFE procedure~\cite{safe2024} for document screening and convergence-based stopping~\cite{bloodgood2009stopping} for active learning, these methods have not been adapted to DNN test selection, where the goal is efficient fault discovery.

A key empirical observation motivates our approach: test selection often exhibits \textit{diminishing returns}. Inputs with higher uncertainty, which are typically selected first, tend to reveal faults at higher rates. As testing progresses, the remaining inputs yield progressively fewer faults per labeled instance.

This behavior suggests that the value of additional labeling decreases over time, creating a natural point at which further testing becomes inefficient. Based on this observation, we propose \textbf{AdaStop}, a cost-aware framework that terminates testing when the marginal fault discovery rate $p(t)$ falls below a cost-adjusted threshold $\tau = c/v$, where $c$ denotes the labeling cost and $v$ represents the value of discovering a fault.

The AdaStop framework consists of three components: (1) an uncertainty-based selection strategy (DeepGini) to prioritize inputs likely to reveal faults, (2) a sliding-window estimator of the marginal fault discovery rate, and (3) a cost-aware stopping rule that terminates labeling once the estimated rate falls below $\tau$.

\textbf{Contributions.} (1)~We frame DNN test selection as a sequential cost-benefit optimization, deriving the optimal stopping condition $\tau = c/v$. (2)~We propose a cost-aware threshold-based stopping rule and compare it against practical alternatives, including patience, consecutive non-faults, confidence-based, and cumulative-rate stopping. (3)~We conduct a comprehensive evaluation across 3 datasets, 4 architectures, 3 quality levels, and 8 strategies, achieving 65--84\% recall with 70--91\% budget savings. (4)~We provide an open-source implementation for community adoption.

\section{Problem Formulation}

\subsection{Setting and Objective}

Consider a DNN classifier $M: \mathcal{X} \rightarrow \mathcal{Y}$ and an unlabeled test pool $\mathcal{U} = \{x_1, \ldots, x_n\}$. A \textit{fault} is an input where $M(x) \neq y^*$ (the ground-truth label). Labeling costs $c > 0$ per query; each fault discovered has value $v > 0$.

Test selection proceeds sequentially: at step $t$, select input $x_t$ from the remaining pool via strategy $s$, query the oracle for label $y_t$, record outcome $r_t = \mathbf{1}[M(x_t) \neq y_t]$, and decide whether to continue. The \textit{net value} at stopping time $T$ is:
\begin{equation}
V(T) = v \cdot F(T) - c \cdot T
\end{equation}
where $F(T) = \sum_{t=1}^{T} r_t$ is the number of faults found.

\subsection{Optimal Stopping Threshold}

At step $t$, the marginal cost of labeling one more input is $c$, and the marginal benefit is $v \cdot p(t)$, where $p(t) = \mathbb{E}[r_t]$ is the probability of discovering a fault. The marginal net value of the next label is:
\begin{equation}
\Delta V(t) = v \cdot p(t) - c
\end{equation}

The optimal decision rule follows directly: \textbf{continue} if $v \cdot p(t) > c$ (positive marginal value); \textbf{stop} if $v \cdot p(t) \leq c$. Rearranging the stopping condition gives:
\begin{equation}
\boxed{\tau = \frac{c}{v}}
\end{equation}

\textbf{Proposition 1 (Optimal Stopping).} Under diminishing returns (fault rate $p(t)$ is non-increasing), the optimal policy stops at the first $T$ where $p(T) \leq \tau = c/v$. Once $p(t)$ drops below $\tau$, it remains below for all subsequent steps, so stopping immediately is optimal. \hfill $\square$

The threshold $\tau$ is the \textit{break-even fault rate}. For example, if labeling costs $c = \$1$ and each fault is worth $v = \$20$, then $\tau = 0.05$: we stop when fewer than 5\% of labels are expected to reveal faults. At this point, the expected value of the next label exactly equals its cost. The threshold automatically adapts---safety-critical scenarios (low $c/v$) continue longer; expensive-labeling scenarios (high $c/v$) stop earlier.

\subsection{Fault Rate Estimation}

Since $p(t)$ is not directly observable, we estimate it from recent history using a sliding window:
\begin{equation}
\hat{p}(t) = \frac{1}{W} \sum_{i=t-W+1}^{t} r_i
\end{equation}
where $W$ is the window size. This estimator addresses three challenges: (1) \textit{non-stationarity}---the fault rate decreases as testing progresses, so we use only recent samples rather than the full history; (2) \textit{noise}---individual outcomes are binary, so averaging over $W$ samples provides smoothing; (3) \textit{responsiveness}---the window adapts to rate changes faster than a cumulative average. The choice of $W$ involves a bias-variance trade-off: small $W$ is responsive but noisy (risking premature stopping); large $W$ is stable but slow to detect rate drops. We analyze this empirically in Section~\ref{sec:results}.

\subsection{The Diminishing Returns Property}

The optimal stopping condition relies on a key empirical observation: \textit{test selection exhibits diminishing returns}. When using uncertainty-based selection (e.g., DeepGini), high-uncertainty inputs are selected first. Since high uncertainty correlates with higher fault probability, the fault rate $p(t)$ decreases as $t$ increases. This ensures a natural stopping point where $p(t)$ crosses below $\tau$. Without this property, the stopping problem would be ill-defined---the rate could fluctuate unpredictably. We validate this assumption formally with Mann-Kendall monotonicity tests in Section~\ref{sec:results}.

\section{Related Work}

\subsection{Test Selection for DNNs}

\textit{Uncertainty-based methods} prioritize inputs where the model is least confident. DeepGini~\cite{feng2020deepgini} uses Gini impurity of softmax outputs; FAST~\cite{chen2024fast} addresses over-confidence through guided feature selection; CertPri~\cite{zheng2023certpri} uses movement cost with formal robustness guarantees. \textit{Coverage and diversity-based methods} aim for diverse test inputs: NLC~\cite{yuan2023nlc} captures neuron output distributions; DeepGD~\cite{aghababaeyan2024deepgd} combines Gini score with geometric diversity via NSGA-II. \textit{Learning-based methods} include TestRank~\cite{li2021testrank} (GNN-based ranking) and ATS~\cite{gao2022ats} (RL-based adaptive selection). \textit{Sampling-based methods} include DeepSample~\cite{guerriero2024deepsample} and DeepReduce~\cite{zhou2020deepreduce}. All these methods use fixed budgets without principled stopping criteria.

\subsection{Stopping Criteria in Other Domains}

Active learning uses convergence-based criteria~\cite{bloodgood2009stopping,ishibashi2020stopping,ishibashi2021error}---stopping when the model stops improving. These are fundamentally different from our setting: we are not training a model, but discovering faults in a fixed model. Document screening uses discovery-based criteria: SAFE~\cite{safe2024} monitors discovery rates; Chao's estimator~\cite{bron2024chao} predicts total relevant documents. Mittal et al.~\cite{mittal2024adaptive} formulate model evaluation as an MDP but minimize estimation error, not fault discovery efficiency. We adapt the discovery-based stopping concept, adding a cost-benefit formulation specific to DNN testing.

\subsection{Research Gap}

Existing methods typically assume a fixed labeling budget and do not provide principled guidance on when additional testing is no longer worthwhile. Table~\ref{tab:gap} summarizes the comparison.

\begin{table}[t]
\centering
\caption{Comparison of Related Work}
\label{tab:gap}
\resizebox{\columnwidth}{!}{%
\begin{tabular}{lcccc}
\toprule
\textbf{Method} & \textbf{Selection} & \textbf{Stopping} & \textbf{Cost} & \textbf{DNN} \\
\midrule
DeepGini~\cite{feng2020deepgini} & Uncertainty & Fixed & No & Yes \\
TestRank~\cite{li2021testrank} & Learned & Fixed & No & Yes \\
ATS~\cite{gao2022ats} & RL-based & Fixed & No & Yes \\
DeepSample~\cite{guerriero2024deepsample} & Sampling & Fixed & No & Yes \\
AL Stopping~\cite{bloodgood2009stopping} & Active L. & Convergence & No & No \\
SAFE~\cite{safe2024} & N/A & Discovery & No & No \\
\textbf{AdaStop (Ours)} & \textbf{Any} & \textbf{Cost-benefit} & \textbf{Yes} & \textbf{Yes} \\
\bottomrule
\end{tabular}%
}
\end{table}

\section{Proposed Approach}

\subsection{Framework Overview}

\begin{figure}[t]
\centering
\begin{tikzpicture}[
    scale=1,
    transform shape,
    node distance=0.5cm and 0.8cm,
    box/.style={rectangle, draw=black!70, fill=blue!8, thick, minimum height=0.8cm, minimum width=2cm, text centered, font=\footnotesize},
    decision/.style={rectangle, draw=red!70, fill=red!8, thick, minimum height=0.9cm, minimum width=4.5cm, text centered, font=\footnotesize},
    input/.style={rectangle, draw=gray!70, fill=gray!10, thick, minimum height=0.6cm, minimum width=1.8cm, text centered, font=\scriptsize},
    arrow/.style={->, thick, >=stealth},
]
\node[input] (pool) {Test Pool $\mathcal{U}$};
\node[input, right=1cm of pool] (model) {Model $M$};
\node[input, right=1cm of model] (params) {$\tau$, $W$, $c$, $v$};
\node[box, below=0.6cm of pool, align=center] (selector) {\textbf{Selection}\\(DeepGini)};
\node[box, below=0.6cm of model, align=center] (oracle) {\textbf{Oracle}\\(Label)};
\node[box, below=0.6cm of params, align=center] (estimator) {\textbf{Rate}\\Estimator};
\node[decision, below=0.8cm of oracle] (stop) {\textbf{Stop when:} $\hat{p}(t) < \tau$};
\node[input, below=0.6cm of stop] (output) {Test Set $\mathcal{L}$};
\draw[arrow] (pool) -- (selector);
\draw[arrow] (model) -- (oracle);
\draw[arrow] (params) -- (estimator);
\draw[arrow] (selector) -- (oracle) node[midway,above,font=\tiny] {$x_t$};
\draw[arrow] (oracle) -- (estimator) node[midway,above,font=\tiny] {$r_t$};
\draw[arrow] (selector.south) -- ++(0,-0.35) -| (stop.north west);
\draw[arrow] (estimator.south) -- ++(0,-0.35) -| (stop.north east);
\draw[arrow] (stop) -- (output);
\draw[arrow,dashed,gray,rounded corners=3pt] (stop.west) -- ++(-2,0) node[pos=0.3,above,font=\tiny] {continue} |- (selector.west);
\end{tikzpicture}
\caption{AdaStop Framework. The system iteratively selects inputs via DeepGini, obtains labels, estimates the fault rate, and stops when $\hat{p}(t) < \tau$.}
\label{fig:framework}
\end{figure}
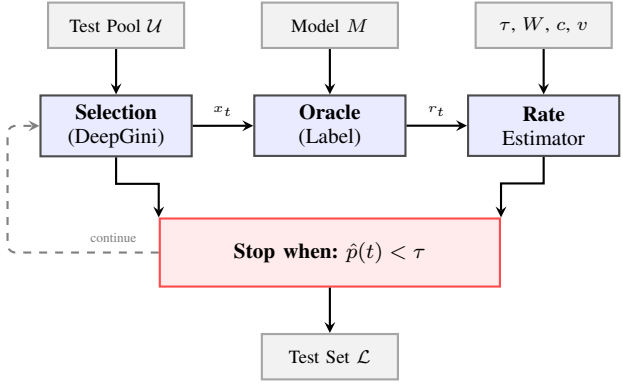

Figure~\ref{fig:framework} illustrates the AdaStop framework. The system iteratively selects inputs via uncertainty-based ranking, obtains labels from an oracle, estimates the current fault rate, and stops when the rate falls below the cost-justified threshold.

\subsection{Selection Strategy}

AdaStop uses DeepGini~\cite{feng2020deepgini} as its default selection strategy, prioritizing inputs by Gini impurity of softmax outputs:
\begin{equation}
\text{Gini}(x) = 1 - \sum_{i=1}^{C} p_i(x)^2
\end{equation}
where $p_i(x)$ is the predicted probability for class $i$ and $C$ is the number of classes. At each step, we select the input with the highest Gini score from the remaining pool, as uncertain predictions are more likely to be incorrect. We choose DeepGini for its simplicity (requires only softmax outputs), effectiveness (4--5$\times$ over random), and wide adoption. AdaStop's stopping framework is \textit{strategy-agnostic} and is compatible with any strategy that produces a ranked ordering (Section~\ref{sec:strategy}).

\subsection{Stopping Criteria}

We discuss four practical stopping criteria for different scenarios:

\textbf{Threshold-based (default).} Stop when $\hat{p}(t) < \tau$, with a minimum sample constraint $N_{\min}$ to ensure reliable estimates. Directly implements the cost-benefit optimality condition. The threshold $\tau = c/v$ should reflect the testing scenario's cost-benefit ratio.

\textbf{Patience-based.} When $\hat{p}(t)$ first drops below $\tau$, continue for an additional patience window of $k$ labels and stop unless the estimated rate rises above $\tau$ again. This criterion adds a small safety margin against transient fluctuations at the stopping boundary.

\textbf{Consecutive non-faults.} Stop after observing $k$ consecutive labels with no faults. This criterion is simple to implement and does not require setting an explicit cost-benefit threshold, but it is less directly tied to the optimality condition.

\textbf{Confidence-based.} For risk-averse scenarios, stop when the upper bound of a Wilson confidence interval~\cite{wilson1927probable} falls below $\tau$: $\text{CI}_{\text{upper}}(\hat{p}) < \tau$. This reduces premature stopping at the cost of more budget.

\subsection{Complete Algorithm}

Algorithm~\ref{alg:adastop} presents the complete AdaStop procedure.

\begin{algorithm}[t]
\caption{AdaStop: Cost-Aware Test Selection}
\label{alg:adastop}
\begin{algorithmic}[1]
\STATE \textbf{Input:} Test pool $\mathcal{U}$, model $M$, threshold $\tau$, window $W$, min samples $N_{\min}$
\STATE \textbf{Output:} Labeled test set $\mathcal{L}$
\STATE Compute Gini scores $G_i \gets 1 - \sum_{j} P_{ij}^2$ for all $x_i \in \mathcal{U}$
\STATE $\mathcal{R} \gets \mathcal{U}$, $\mathcal{L} \gets \emptyset$, $H \gets []$
\FOR{$t = 1, 2, \ldots$}
\STATE $x_t \gets \arg\max_{x_i \in \mathcal{R}} G_i$; \quad $\mathcal{R} \gets \mathcal{R} \setminus \{x_t\}$
\STATE $r_t \gets \mathbf{1}[M(x_t) \neq \textsc{Oracle}(x_t)]$; \quad append $r_t$ to $H$
\STATE $\mathcal{L} \gets \mathcal{L} \cup \{(x_t, r_t)\}$
\STATE $\hat{p} \gets \text{mean}(H[\max(1,|H|-W+1):|H|])$
\IF{$|\mathcal{L}| \geq N_{\min}$ \AND $\hat{p} < \tau$}
\RETURN $\mathcal{L}$ \COMMENT{Cost-aware stop}
\ENDIF
\ENDFOR
\end{algorithmic}
\end{algorithm}

\section{Experimental Setup}

\subsection{Research Questions}

\begin{itemize}
    \item \textbf{RQ1}: Can AdaStop realize high recall with a small budget fraction?
    \item \textbf{RQ2}: What is the impact of $\tau$ on recall-budget balance?
    \item \textbf{RQ3}: What are the differences between various stopping criteria? 
    \item \textbf{RQ4}: How sensitive is AdaStop to window size $W$?
    \item \textbf{RQ5}: Does AdaStop have good generalization properties with respect to datasets and architectures?
    \item \textbf{RQ6}: Is AdaStop strategy-agnostic?
    \item \textbf{RQ7}: How sensitive is AdaStop to model quality?
\end{itemize}

\subsection{Datasets, Model Architectures, and Configuration}

We evaluate across three datasets: \textbf{CIFAR-10}~\cite{krizhevsky2009cifar} (10K test images), \textbf{SVHN}~\cite{netzer2011svhn} (26K test images), and \textbf{FashionMNIST}~\cite{xiao2017fashion} (10K test images); four architectures: \textbf{ResNet-20}~\cite{he2016resnet}, \textbf{VGG-16}~\cite{simonyan2015vgg}, \textbf{DenseNet-121}~\cite{huang2017densenet}, and \textbf{ShuffleNetV2}~\cite{ma2018shufflenetv2}; and three model quality levels for ResNet-20 on CIFAR-10: low ($\sim$70\%, 2,961 faults), mid ($\sim$79\%, 2,117 faults), and high ($\sim$88\%, 1,196 faults). Table~\ref{tab:all_configs} summarizes the configurations.

\begin{table}[t]
\centering
\caption{Dataset--Architecture Configurations}
\label{tab:all_configs}
\small
\begin{tabular}{llrr}
\toprule
\textbf{Dataset} & \textbf{Architecture} & \textbf{Test Pool} & \textbf{Faults} \\
\midrule
CIFAR-10 & ResNet-20 & 10,000 & 1,196 \\
CIFAR-10 & VGG-16 & 10,000 & 811 \\
CIFAR-10 & DenseNet-121 & 10,000 & 1,494 \\
CIFAR-10 & ShuffleNetV2 & 10,000 & 1,427 \\
SVHN & ResNet-20 & 26,032 & 1,159 \\
FashionMNIST & ResNet-20 & 10,000 & 570 \\
\bottomrule
\end{tabular}
\end{table}

\textbf{Default parameters:} $\tau{=}0.05$, $W{=}20$, $N_{\min}{=}50$. We compare against DeepGini at fixed budgets ($k \in \{1,2,5,10,20,50,100\}\%$), an oracle (perfect fault-first ordering), and five alternative stopping criteria: patience-5, consecutive non-faults ($k{=}50,100$), confidence-90\%, and cumulative rate. For RQ6, we test eight selection strategies: DeepGini, entropy, margin, boundary distance, random, TestRank~\cite{li2021testrank}, ATS~\cite{gao2022ats}, and DeepGD~\cite{aghababaeyan2024deepgd}.

\textbf{Metrics:} (1)~\textbf{Budget used (\%):} fraction of the test pool labeled before stopping. (2)~\textbf{Fault recall (\%):} fraction of total faults discovered. (3)~\textbf{Efficiency:} ratio of faults found to labels used. (4)~\textbf{Net value:} $v \times \text{faults} - c \times \text{budget}$ with $v{=}20$, $c{=}1$.

\subsection{Experimental Procedure}

For each configuration, we compute model predictions and Gini scores, sort inputs by score, iterate through the ranked list while updating the sliding-window estimate, and apply the stopping rule once $N_{\min}$ samples have been collected. For baseline comparisons, we run DeepGini at each fixed budget and record the same metrics.

\section{Experimental Results}
\label{sec:results}

\subsection{RQ1: Effectiveness}

Table~\ref{tab:results} compares AdaStop against exhaustive testing on CIFAR-10/ResNet-20.

\begin{table}[t]
\centering
\caption{AdaStop vs.\ Exhaustive Testing ($\tau{=}0.05$, CIFAR-10)}
\label{tab:results}
\begin{tabular}{lccc}
\toprule
\textbf{Metric} & \textbf{Full (100\%)} & \textbf{AdaStop} & \textbf{Improv.} \\
\midrule
Budget Used & 100\% & 23.5\% & 76.5\% saved \\
Faults Found & 1,196 (100\%) & 945 (79.0\%) & -- \\
Efficiency & 0.120 & 0.402 & 3.35$\times$ \\
Net Value & 13,920 & 16,552 & +18.9\% \\
\bottomrule
\end{tabular}
\end{table}

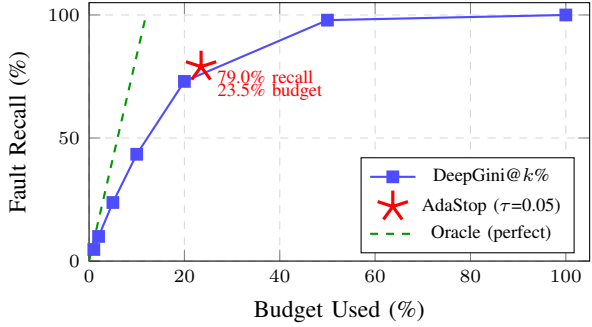
\begin{figure}[t]
\centering
\begin{tikzpicture}
\begin{axis}[
    width=0.95\columnwidth,
    height=5cm,
    xlabel={Budget Used (\%)},
    ylabel={Fault Recall (\%)},
    xmin=0, xmax=105,
    ymin=0, ymax=105,
    grid=major,
    grid style={dashed,gray!30},
    legend pos=south east,
    legend style={font=\scriptsize},
    tick label style={font=\scriptsize},
    label style={font=\small},
]
\addplot[color=blue!70, mark=square*, mark size=2pt, thick] coordinates {
    (1, 4.7) (2, 10.0) (5, 23.8) (10, 43.4) (20, 73.0) (50, 97.9) (100, 100)
};
\addlegendentry{DeepGini@$k$\%}
\addplot[only marks, mark=star, mark size=6pt, color=red, very thick] coordinates {(23.5, 79.0)};
\addlegendentry{AdaStop ($\tau$=0.05)}
\addplot[color=green!60!black, dashed, thick, domain=0:11.96] {x * 8.36};
\addlegendentry{Oracle (perfect)}
\node[anchor=west, font=\scriptsize, text=red] at (axis cs:25,75) {79.0\% recall};
\node[anchor=west, font=\scriptsize, text=red] at (axis cs:25,69) {23.5\% budget};
\end{axis}
\end{tikzpicture}
\caption{Budget-recall trade-off. AdaStop (star) identifies stopping points close to the optimal point on the budget–recall curve.}
\label{fig:pareto}
\end{figure}

Figure~\ref{fig:pareto} shows the budget-recall trade-off. AdaStop automatically finds an effective operating point on the budget-recall curve without manual budget tuning.

\begin{figure}[t]
\centering
\begin{tikzpicture}
\begin{axis}[
    width=0.95\columnwidth,
    height=5cm,
    xlabel={Budget Used (\%)},
    ylabel={Net Value ($v \cdot \text{faults} - c \cdot \text{budget}$)},
    xmin=0, xmax=105,
    ymin=0, ymax=25000,
    grid=major,
    grid style={dashed,gray!30},
    legend pos=south east,
    legend style={font=\scriptsize},
    tick label style={font=\scriptsize},
    label style={font=\small},
    scaled y ticks=false,
    yticklabel style={/pgf/number format/fixed, /pgf/number format/1000 sep={\,}},
]
\addplot[color=green!60!black, mark=diamond*, mark size=2pt, thick, dashed] coordinates {
    (1, 1900) (2, 3800) (5, 9500) (10, 19000) (11.96, 22724) (20, 21920) (50, 18920) (100, 13920)
};
\addlegendentry{Oracle (perfect)}
\addplot[color=blue!70, mark=square*, mark size=2pt, thick] coordinates {
    (1, 1020) (2, 2180) (5, 5200) (10, 9380) (20, 15460) (50, 18420) (100, 13920)
};
\addlegendentry{DeepGini@$k$\%}
\addplot[only marks, mark=star, mark size=6pt, color=red, very thick] coordinates {(23.5, 16552)};
\addlegendentry{AdaStop ($\tau$=0.05)}
\node[anchor=south, font=\scriptsize, text=green!50!black] at (axis cs:11.96,22724) {Oracle peak};
\node[anchor=south west, font=\scriptsize, text=red] at (axis cs:25,16552) {16,552};
\end{axis}
\end{tikzpicture}
\caption{Net value comparison ($v{=}20$, $c{=}1$). Oracle peaks at 11.96\% budget. DeepGini peaks at 50\% then decreases. Exhaustive testing (100\%) yields the \textit{lowest} net value among non-trivial budgets.}
\label{fig:netvalue}
\end{figure}
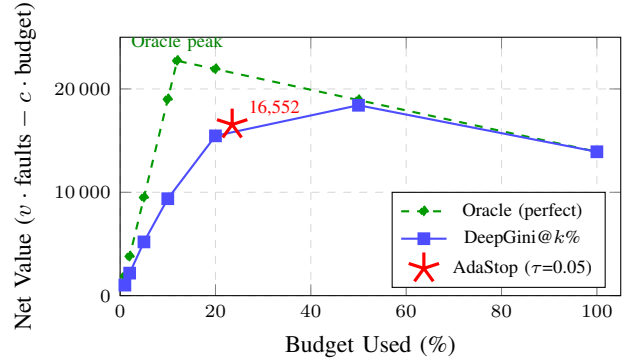

Figure~\ref{fig:netvalue} compares net value across methods. The oracle achieves peak net value at exactly 11.96\% budget (the error rate), where all faults are found with zero waste. DeepGini peaks at 50\% and then decreases as costs outpace discovery. AdaStop achieves 90\% of DeepGini's peak value while using 27 percentage points less budget. Notably, exhaustive testing yields \textit{lower} net value (13,920) than AdaStop (16,552), demonstrating that more testing is not always better.

\textbf{RQ1:} AdaStop achieves 79.0\% fault recall using only 23.5\% of the budget, with a 3.35$\times$ efficiency improvement and 18.9\% higher net value than exhaustive testing.

\subsection{RQ2: Threshold Selection}

Table~\ref{tab:threshold} shows threshold sensitivity.

\begin{table}[t]
\centering
\caption{Threshold Selection Results (CIFAR-10, ResNet-20)}
\label{tab:threshold}
\small
\begin{tabular}{lccc}
\toprule
\textbf{Threshold $\tau$} & \textbf{Budget} & \textbf{Recall} & \textbf{Efficiency} \\
\midrule
0.01--0.05 & 23.5\% & 79.0\% & 0.402 \\
0.10 & 13.3\% & 54.7\% & 0.494 \\
0.20 & 13.2\% & 54.6\% & 0.495 \\
\bottomrule
\end{tabular}
\end{table}

\textbf{RQ2:} Thresholds $\tau{=}0.01$--$0.05$ converge to the same stopping point (23.5\% budget, 79.0\% recall), while $\tau{=}0.10$--$0.20$ converge to an earlier point (13.2\% budget, 54.7\% recall). This stepped behavior indicates the fault rate exhibits sharp drops rather than smooth decline, creating natural stopping plateaus where multiple threshold values trigger at similar points.

\subsection{RQ3: Stopping Criteria Comparison}

Table~\ref{tab:criteria} compares stopping criteria, all using DeepGini for selection.

\begin{table}[t]
\centering
\caption{Stopping Criteria Comparison (CIFAR-10, ResNet-20)}
\label{tab:criteria}
\small
\begin{tabular}{lrrrr}
\toprule
\textbf{Criterion} & \textbf{Budget} & \textbf{Recall} & \textbf{Effic.} & \textbf{Net Val.} \\
\midrule
Threshold & 23.5\% & 79.0\% & 0.403 & 16,552 \\
Patience-5 & 27.6\% & 85.4\% & 0.370 & 17,659 \\
Consec.-50 & 32.9\% & 90.1\% & 0.328 & 18,272 \\
Consec.-100 & 56.1\% & 98.8\% & 0.211 & 18,015 \\
Confidence-90 & 60.3\% & 99.0\% & 0.196 & 17,647 \\
Cumul.\ rate & 100.0\% & 100.0\% & 0.120 & 13,920 \\
\bottomrule
\end{tabular}
\end{table}

The criteria form a clear aggressive-to-conservative spectrum. Threshold is most efficient; patience-5 adds a small safety margin; consecutive-50 achieves the best net value; and consecutive-100 and confidence-90 are conservative, achieving $\geq$98.8\% recall at higher cost. The cumulative rate criterion never triggers because the cumulative average is too slow to reflect marginal rate decline.

\textbf{RQ3:} For maximum efficiency, use threshold-based stopping. For best net value, consecutive-50 is optimal. For near-complete recall, confidence-90 provides 99.0\% recall with 40\% savings.

\subsection{RQ4: Sensitivity to Window Size}

\begin{table}[t]
\centering
\caption{Window Size Sensitivity ($\tau{=}0.05$, CIFAR-10)}
\label{tab:window}
\small
\begin{tabular}{lrrrl}
\toprule
\textbf{$W$} & \textbf{Budget} & \textbf{Recall} & \textbf{Effic.} & \textbf{Status} \\
\midrule
10 & 6.5\% & 30.1\% & 0.551 & Too aggressive \\
20 & 23.5\% & 79.0\% & 0.402 & Recommended \\
50 & 23.5\% & 79.0\% & 0.403 & Stable \\
100 & 33.2\% & 90.2\% & 0.325 & Conservative \\
\bottomrule
\end{tabular}
\end{table}

Small windows ($W{=}10$) cause premature stopping; $W{=}20$ and $W{=}50$ converge to the same point, showing robustness for moderate sizes. The minimum-samples parameter $N_{\min}$ is also robust: values of 20, 50, and 100 all produce identical results because the natural stopping point (2,348 samples) far exceeds $N_{\min}$.

\subsection{RQ5: Generalization}

\subsubsection{Cross-Dataset}

AdaStop consistently achieves substantial budget savings (77--91\%) across all datasets while maintaining 65--79\% recall. 

\begin{table}[t]
\centering
\caption{Cross-Dataset Results (ResNet-20, $\tau{=}0.05$)}
\label{tab:multi_dataset}
\small
\begin{tabular}{lrrrr}
\toprule
\textbf{Dataset} & \textbf{Budget} & \textbf{Recall} & \textbf{Savings} & \textbf{Effic.} \\
\midrule
CIFAR-10 & 23.5\% & 79.0\% & 76.5\% & 0.403 \\
SVHN & 8.6\% & 65.1\% & 91.4\% & 0.338 \\
FashionMNIST & 12.4\% & 71.4\% & 87.6\% & 0.327 \\
\midrule
\textit{Average} & \textit{14.8\%} & \textit{71.8\%} & \textit{85.2\%} & \textit{0.356} \\
\bottomrule
\end{tabular}
\end{table}

\subsubsection{Cross-Architecture}

\begin{table}[t]
\centering
\caption{Cross-Architecture Results (CIFAR-10, $\tau{=}0.05$)}
\label{tab:multi_arch}
\small
\resizebox{\columnwidth}{!}{%
\begin{tabular}{lrrrrr}
\toprule
\textbf{Architecture} & \textbf{Faults} & \textbf{Budget} & \textbf{Recall} & \textbf{Savings} & \textbf{Effic.} \\
\midrule
ResNet-20 & 1,196 & 23.5\% & 79.0\% & 76.5\% & 0.403 \\
VGG-16 & 811 & 18.0\% & 80.0\% & 82.0\% & 0.360 \\
DenseNet-121 & 1,494 & 21.3\% & 66.6\% & 78.7\% & 0.468 \\
ShuffleNetV2 & 1,427 & 30.3\% & 84.0\% & 69.7\% & 0.396 \\
\midrule
\textit{Average} & --- & \textit{23.3\%} & \textit{77.4\%} & \textit{76.7\%} & \textit{0.407} \\
\bottomrule
\end{tabular}%
}
\end{table}

Across architectures, the 18--30\% budget range and the 67--84\% recall range confirm that the stopping criterion adapts to varying error rates without parameter changes.

\textbf{RQ5:} AdaStop generalizes well: 65--84\% recall with 70--91\% savings across 3 datasets and 4 architectures.

\subsection{RQ6: Strategy-Agnostic Property}
\label{sec:strategy}

\begin{table}[t]
\centering
\caption{Strategy-Agnostic Evaluation (CIFAR-10, $\tau{=}0.05$)}
\label{tab:strategy}
\small
\resizebox{\columnwidth}{!}{%
\begin{tabular}{lrrrr}
\toprule
\textbf{Strategy} & \textbf{Budget} & \textbf{Recall} & \textbf{Savings} & \textbf{Effic.} \\
\midrule
DeepGini & 23.5\% & 79.0\% & 76.5\% & 0.403 \\
Entropy & 24.2\% & 80.7\% & 75.8\% & 0.399 \\
Margin & 24.4\% & 80.4\% & 75.6\% & 0.394 \\
Boundary & 24.4\% & 80.4\% & 75.6\% & 0.394 \\
TestRank & 23.5\% & 79.0\% & 76.5\% & 0.402 \\
ATS & 23.8\% & 79.2\% & 76.2\% & 0.398 \\
DeepGD & 12.0\% & 43.5\% & 88.0\% & 0.432 \\
Random & 1.1$\pm$0.8\% & 1.2$\pm$0.9\% & 98.9\% & 0.113 \\
\bottomrule
\end{tabular}%
}
\end{table}

Six of the eight strategies produce remarkably consistent stopping behavior: 23--24\% budget, 79--81\% recall, and 0.39--0.40 efficiency. This consistency occurs because all six strategies exploit model uncertainty to order inputs, producing similar diminishing-returns curves that cross $\tau$ at nearly the same point.

\textbf{RQ6:} AdaStop is strategy-agnostic---all uncertainty-based strategies produce consistent stopping points. DeepGD stops earlier (12\%) due to a steeper initial rate decline from its diversity component; random stops almost immediately because its unprioritized fault rate fluctuates enough to cross $\tau$ soon after the initial window.

\subsection{RQ7: Effect of Model Quality}

\begin{table}[t]
\centering
\caption{Model Quality Sensitivity (CIFAR-10, ResNet-20)}
\label{tab:quality}
\small
\resizebox{\columnwidth}{!}{%
\begin{tabular}{lrrrrr}
\toprule
\textbf{Quality} & \textbf{Faults} & \textbf{Budget} & \textbf{Recall} & \textbf{Savings} & \textbf{Effic.} \\
\midrule
Low ($\sim$70\%) & 2,961 & 64.7\% & 93.7\% & 35.3\% & 0.429 \\
Mid ($\sim$79\%) & 2,117 & 39.3\% & 80.7\% & 60.7\% & 0.435 \\
High ($\sim$88\%) & 1,196 & 23.5\% & 79.0\% & 76.5\% & 0.403 \\
\bottomrule
\end{tabular}%
}
\end{table}

As model quality improves (fewer faults), AdaStop uses less budget and achieves higher savings. For the low-quality model (30\% error rate), the fault rate remains above $\tau$ for longer, while for the high-quality model (12\% error rate), it drops below $\tau$ at 23.5\%. The stopping point therefore varies naturally with fault density.

\textbf{RQ7:} AdaStop adapts to model quality: more budget for worse models (64.7\%) and less for better ones (23.5\%), with consistent efficiency ($\sim$0.40--0.44).

\subsection{Diminishing Returns Validation}

The diminishing returns property is validated via APFD scores and Mann-Kendall monotonicity tests (Table~\ref{tab:validation}).

\begin{table}[t]
\centering
\caption{Diminishing Returns: APFD and Mann-Kendall Tests}
\label{tab:validation}
\small
\resizebox{\columnwidth}{!}{%
\begin{tabular}{llccrl}
\toprule
\textbf{Dataset} & \textbf{Strategy} & \textbf{APFD} & \textbf{MK $\tau$} & \textbf{$p$-val} & \textbf{Trend} \\
\midrule
CIFAR-10 & DeepGini & 0.849 & $-0.727$ & $<10^{-4}$ & Decreasing \\
CIFAR-10 & Entropy & 0.850 & $-0.744$ & $<10^{-4}$ & Decreasing \\
CIFAR-10 & Random & 0.502 & $-0.038$ & $<10^{-4}$ & No trend \\
SVHN & DeepGini & 0.885 & $-0.512$ & $<10^{-4}$ & Decreasing \\
FashionMNIST & DeepGini & 0.901 & $-0.631$ & $<10^{-4}$ & Decreasing \\
\bottomrule
\end{tabular}%
}
\end{table}

All guided strategies show strong decreasing trends (MK $\tau$ from $-0.51$ to $-0.74$, all $p < 10^{-4}$), confirming the diminishing returns assumption. Random selection shows near-zero trend ($|\tau| < 0.04$), confirming no meaningful trend as expected---without prioritization, the fault rate remains roughly constant at the base error rate.

\subsection{Summary of Key Findings}

\begin{enumerate}
    \item \textbf{Cost-aware stopping improves labeling efficiency}: AdaStop with $\tau{=}0.05$ achieves 79.0\% recall using 23.5\% of the budget---a 3.35$\times$ efficiency gain with 18.9\% higher net value than exhaustive testing.
    \item \textbf{Threshold exhibits stepped behavior}: $\tau{=}0.01$--$0.05$ converge to 23.5\% budget; $\tau{=}0.10$--$0.20$ converge to 13.2\%, reflecting sharp fault rate drops.
    \item \textbf{Multiple criteria suit different needs}: threshold for efficiency, consecutive-50 for best net value, confidence-90 for near-complete recall.
    \item \textbf{Window size matters}: $W \geq 20$ is required for stable estimates; $W{=}10$ causes premature stopping (30\% recall vs.\ 79\%).
    \item \textbf{Generalizes broadly}: consistent 65--84\% recall with 70--91\% savings across 3 datasets, 4 architectures, and 8 strategies.
    \item \textbf{Adapts to model quality}: uses 23--65\% budget depending on error rate, without parameter changes.
\end{enumerate}

\subsection{Limitations}

Our evaluation focuses on image classification across three vision datasets; generalization to NLP, tabular, or multi-modal domains remains to be validated. AdaStop assumes labels are obtained one at a time; batch labeling scenarios (where multiple inputs are sent for annotation simultaneously) would require extending the framework. The threshold $\tau = c/v$ requires practitioners to estimate the labeling cost and fault value; in practice these may be uncertain, though the relative and consecutive stopping criteria partially address this by not requiring explicit cost-benefit ratios. Finally, while the diminishing returns assumption is validated empirically via Mann-Kendall tests across three datasets, it may not hold for models with multi-modal failure clusters or adversarial inputs where faults appear in unexpected regions of the uncertainty space.

\section{Conclusion}

We presented AdaStop, a cost-aware framework for determining when to stop labeling in DNN test selection. By formalizing the stopping problem as a cost-benefit optimization, we derived the optimal stopping threshold $\tau = c/v$ and implemented it through a sliding window fault rate estimator. Our comprehensive evaluation across 3 datasets (CIFAR-10, SVHN, FashionMNIST), 4 architectures (ResNet-20, VGG-16, DenseNet-121, ShuffleNetV2), 3 model quality levels, and 8 selection strategies demonstrates that AdaStop consistently achieves 65--84\% fault recall with 70--91\% budget savings. The framework is strategy-agnostic, producing consistent stopping behavior regardless of the underlying selection method. Among the stopping criteria evaluated, consecutive non-faults ($k{=}50$) achieves the best net value, while threshold-based stopping offers the highest efficiency. The diminishing returns assumption underlying our approach is formally validated via Mann-Kendall tests across all configurations. Future work will extend AdaStop to non-vision domains (NLP, tabular data), batch labeling scenarios.

\bibliographystyle{IEEEtran}

\end{document}